# Personality Engineering with AI Agents:
# A New Methodology for Negotiation Research

Michelle A. Vaccaro[a] and Jared R. Curhan[b*]
Sep 7, 2026

**Abstract**

According to canonical negotiation theory, people's success in a negotiation depends on how well they balance competing demands—empathizing and asserting, demonstrating concern for other and concern for self, being soft on the people and hard on the problem. Yet people struggle to manage these tensions, so researchers have lacked the ability to rigorously test the field's prescriptions under controlled conditions. AI agents do not face the same limitations, and their precision, repertoire, consistency, and scalability enable a new class of experiments to contribute to negotiation theory. In this article, we introduce personality engineering—a methodology that precisely parameterizes the personas of AI agents based on established personality frameworks—and propose how this method can be used to study the traits and behaviors of human negotiators. Given its relevance to negotiation theory, we propose the interpersonal circumplex—and its two core dimensions of warmth and dominance—as a foundational coordinate system for the field. This approach offers both a rigorous methodology for testing classic negotiation theories and a practical guide for designing the personas of AI negotiation agents.

---

a Harvard Business School
b MIT Sloan School of Management
* corresponding author (curhan@mit.edu)

## Introduction

The seminal theories of negotiation share a common structural element: tension. Across classic frameworks, negotiation scholars prescribe not a single strategy but a balancing act that requires people to manage seemingly competing demands. They recommend being hard on the problem yet soft on the people [1], maintaining high concern for both self and other [2], and sustaining empathy alongside assertiveness [3]. In practice, however, human negotiators often perceive a tradeoff and behave as if they must choose between competing demands [1,3,4,5,6].

Human negotiators are also imprecise instruments for testing these theories—they interpret the same instructions differently, shift their behavior in response to their counterpart's actions, and are difficult to recruit and coordinate at scale [7,8,9]. As a result, the field's central prescriptions have been tested primarily through coarse categorical comparisons with surprisingly fragmented results [10,11,12,13].

In this article, we propose a methodology that overcomes these constraints by leveraging the unique properties of AI agents. We introduce *personality engineering*—the systematic parameterization of AI agent personas along theoretically grounded dimensions to test negotiation theories and optimize negotiation outcomes in AI-AI and human-AI negotiations. We use the interpersonal circumplex [14,15] as our coordinate system, parameterizing AI agents along its two core dimensions—warmth and dominance—which we map onto the dual imperatives that negotiation theory has long prescribed but struggled to test [1,2,3]. We describe how, unlike human negotiators, AI agents can be positioned anywhere in the warmth-dominance space (see Figure 1), varied in fine-grained increments, and held steady across interactions at scale. We then lay out a research agenda for using this capability to test foundational questions the field has raised but not yet resolved.

## A Coordinate System for Personality Engineering

To move from prescription to experimentation, we need a shared coordinate system that can operationalize the dual imperatives running through negotiation theory. We propose the interpersonal circumplex—a well-established framework from personality and social psychology that organizes interpersonal behavior along two continuous dimensions: warmth and dominance [14,15]. The interpersonal circumplex has a long history of use in personality assessment, clinical psychology, and social interaction research, providing a common language for describing how people relate to one another across contexts [16,17,18,19,20,21].

The central negotiation constructs also map naturally onto these two dimensions [1,2,3]. Concern for other, empathy, and softness on the people correspond to warmth—the degree to which a negotiator attends to, values, and nurtures the counterpart and the relationship. Concern for self, assertiveness, and hardness on the problem correspond to dominance—the degree to which a negotiator advocates for, prioritizes, and advances their own interests and positions. We do not claim that these negotiation theories completely reduce to warmth and dominance, but that the interpersonal circumplex offers a powerful operationalization: its dimensions are continuous rather than categorical, independently measurable, and grounded in decades of interpersonal theory. These properties make the circumplex especially well suited to the kind of precise, parametric variation that the field needs but has so far lacked the means to achieve.

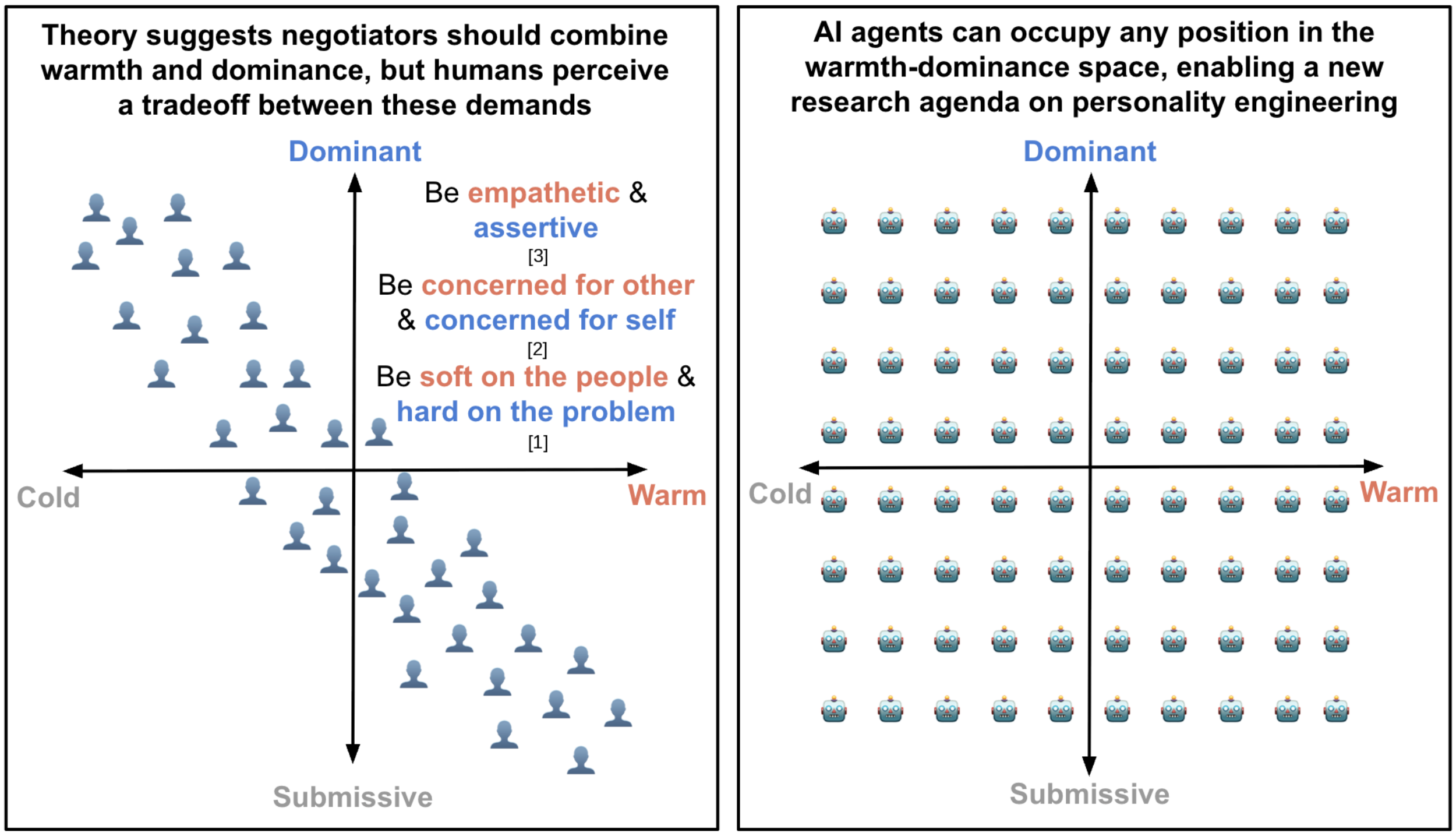

**Figure 1. The interpersonal circumplex as a coordinate system for personality engineering.** The left panel illustrates the gap between theory and practice in negotiations with humans: negotiation theory has long prescribed that effective negotiators must integrate seemingly competing interpersonal demands, yet human negotiators tend to perceive those demands as contradictory, making it difficult for researchers to empirically evaluate the field's core prescriptions. The right panel illustrates the opportunity that AI agents provide: their precision, repertoire, consistency, and scalability enable a new research agenda of personality engineering, the systematic manipulation of AI agent personas to evaluate and advance negotiation theory.

**An Engineering Approach to Studying Negotiation**

With this coordinate system in place, we can now define the methodology it enables: personality engineering [22]. We argue that AI agents make possible a new program of research on personality engineering in negotiation: the deliberate calibration and variation of AI agent interpersonal personas along established theoretical dimensions to evaluate negotiation prescriptions and improve negotiation performance. The approach applies the logic of engineering—specify design variables, define objective functions, optimize outcomes—to the interpersonal dimensions that social psychology has long used to characterize human interaction [23].

In our interpersonal circumplex coordinate system, the primary design variables are the warmth and dominance levels of AI agents, which researchers can specify exogenously on continuous scales. The objective function captures the negotiation outcomes that researchers aim to maximize—such as individual value claimed, joint value created, and counterpart subjective value—with the weights reflecting which outcomes matter most for a given theoretical question [7,24,25].

Given these design variables and objective functions, researchers can optimize AI agent personas by systematically exploring the warmth-dominance space. That is, they can vary warmth and dominance independently, in combination, and at fine-grained intervals to map how each region of the space relates to each outcome of interest with a high degree of control that human studies cannot achieve. What makes this approach feasible are four properties of AI agents: precision, repertoire, consistency, and scalability.

**Precision**

Researchers can specify and manipulate interpersonal traits more precisely with AI agents than with human participants. Instructing a person to be "somewhat warm and somewhat dominant" introduces substantial variation in interpretation and execution, which can confound the dimensions the researcher seeks to isolate [26]. With AI agents, by contrast, warmth and dominance can be assigned target levels—e.g., warmth = 50/100 and dominance = 50/100—and varied through prompting, in-context exemplars, fine-tuning, or architectural interventions. Researchers can also verify them through psychometric instruments administered to the agents [27,28], human ratings of agent behavior [29,30], and analyses of the agent communication [30] to evaluate the extent to which agents maintain such personas across repeated interactions, negotiation contexts, and different counterparts. Together, this capacity for fine-grained parameterization and empirical validation allows negotiation research to move beyond coarse contrasts between behavioral types toward the kind of precise analysis that the field's theories have implied but lacked the means to empirically evaluate.

**Repertoire**

AI agents can occupy any region of the warmth-dominance space, including regions that few human negotiators can reach and sustain. The combination of very high warmth and very high dominance—precisely the profile that many negotiation theories identify as optimal—is rarely achieved among people, and difficult to sustain even when explicitly instructed [6]. AI agents, by contrast, can reliably take on these personas, as well as more extreme or unconventional combinations [30,31,32]. This expanded repertoire allows researchers to examine theoretically important but empirically scarce interpersonal configurations, and to test how negotiation unfolds across all parts of the warmth-dominance space, not just the subset that human participants can comfortably produce. Whether counterparts perceive these scarce configurations as exceptionally skilled or as unnatural is itself an open empirical question, which our methodology is particularly suited to address.

**Consistency**

Additionally, AI agents can better maintain their assigned persona profile across an entire negotiation as they are less affected by the counterpart's behavior. Human negotiators often shift in response to counterpart behavior, fatigue, or emotion, compromising the intended manipulation [9,34,35]. Because AI agents do not have these emotional or cognitive limitations, they can remain warm even when faced with hostility, or remain dominant even when faced with emotional appeals [33]. There are also no spillover effects between negotiations: an AI agent that just faced a hostile counterpart can begin the next interaction with the same profile it started with, uncontaminated by the prior exchange. This consistency helps isolate the effect of the persona configuration itself rather than uncontrolled variation in how faithfully participants enacted it.

**Scalability**

AI agents also enable experiments at a scale that is difficult, if not impossible, to achieve with human participants [36,37]. Researchers can deploy thousands of agents with systematically varied warmth-dominance configurations to negotiate in parallel, generating large datasets that cover the interpersonal space with high resolution [30]. A design that would take months of recruitment, scheduling, and data collection with human participants can run in hours with AI agents [38,39]. This scalability transforms the warmth-dominance space from a theoretical abstraction into an empirically mappable landscape, making it feasible to detect non-linearities, identify interaction effects, and estimate optimal configurations at a resolution that would be prohibitively costly and time-consuming with human samples.

These four properties of AI agents—precision, repertoire, consistency, and scalability—fill the methodological gaps identified earlier in empirical studies with human participants. Researchers can now engineer negotiators who enact the configurations that theory specifies and observe the consequences at scale.

**Research Agenda for Personality Engineering in Negotiation**

We propose six lines of inquiry, each targeting a foundational question, that use this personality engineering approach to address questions negotiation theory raises but existing methods have struggled to answer.

**Optimality**

The most fundamental question personality engineering can address is: what warmth-dominance configuration (or more generally which persona profile) produces the best negotiation outcomes? Existing empirical work offers suggestive but incomplete answers. For example, Jeong, Minson, Yeomans, and Gino [11] found that communicating with warmth in distributive negotiations was counterproductive for economic outcomes, while Kopelman, Rosette, and Thompson [12] found that strategic displays of positive emotion increased the likelihood of closing deals and gaining concessions.

These seemingly contradictory results hint at a complex outcome surface, but each study examines only a slice of the warmth-dominance space using coarse categorical comparisons. Personality engineering enables researchers to map the full surface: to identify not just whether warmth or dominance helps or hurts, but at what levels, in what combinations, and for which outcomes. A configuration that maximizes individual value claimed may differ sharply from one that maximizes joint value created, and both may diverge from the configuration that maximizes counterpart subjective value. Tracing these outcome surfaces—and where the Pareto frontier across competing objectives lies—is an important empirical question that the field's existing methods have not been positioned to answer.

**Contextual Effects**

Importantly, the optimal warmth-dominance configuration likely depends on structural features of the negotiation. Existing research already points to important contextual moderators. Barry and Friedman [10] found that agreeableness and extraversion—constructs closely related to warmth—functioned as liabilities in distributive bargaining but not in integrative contexts. Sinaceur and Tiedens [40] showed that expressing anger—a dominance-related signal—elicited concessions, but only when the counterpart had poor alternatives, suggesting that the returns to dominance depend on the power dynamics of the negotiation.

Personality engineering allows researchers to cross warmth-dominance configurations with these contextual variables in full factorial designs—simultaneously varying negotiation structure (distributive vs. integrative) and power (high vs. low), for example. This makes it possible to test not just whether context moderates the effect of persona on outcomes, but precisely how the optimal configuration shifts across conditions.

**Dyadic Effects**

Negotiation is inherently dyadic, so the effectiveness of any persona configuration also likely depends on who is sitting across the table. Interpersonal theory predicts certain complementarities: warmth invites warmth, dominance invites submissiveness, submissiveness invites dominance [20,41,42]. If these dynamics hold in negotiation, a warm and dominant agent may produce very different outcomes depending on whether it faces a warm and submissive counterpart or a cold and dominant one. In line with this hypothesis, research on individual differences in negotiation confirms that outcomes depend not just on each negotiator's characteristics but on the dyadic interaction between them [43].

Existing studies, however, have lacked the means to systematically and exogenously cross one party's persona with the other's. Personality engineering makes full factorial dyadic designs feasible: researchers can pair agents at every combination of warmth and dominance levels and test when complementarity effects exist—where contrasting profiles produce the best joint outcomes—or when similarity effects prevail, with matching profiles generating greater rapport and cooperation.

**Temporal and Sequential Effects**

In many real-world situations, negotiations unfold over time and the effects of warmth and dominance likely unfold dynamically over the course of an interaction and across repeated encounters. Research on negotiation dynamics suggests that early and late phases of negotiation serve different functions: early exchanges establish rapport, set anchors, and signal intentions, while later exchanges involve concession-making, problem-solving, and closing [9,44,45,46]. Warmth may therefore pay its largest dividends early, when trust is being established and information channels are opening, while dominance may matter most in later phases, when parties must commit to positions and close the deal. In this vein, empirical research shows that subjective value from one negotiation predicts better economic outcomes in the next [47], though it can also trigger hubristic pride that undermines subsequent performance [48]. The sequence of warmth and dominance may also matter independently: other research highlights how negotiators who shift from positive to negative emotion extract larger concessions than those who display negative emotion throughout [49].

Personality engineering enables researchers to investigate these temporal questions with unprecedented control. Does warmth yield little immediate economic advantage but compound across rounds as trust accumulates? Does dominance produce early gains that erode as counterparts grow resentful or disengage? Researchers can also test switching effects—whether an agent that begins with high warmth and shifts to high dominance outperforms one that maintains a fixed profile, and whether the reverse sequence produces different results. These designs require the kind of precise, repeatable control over persona that only AI agents can provide, as human negotiators cannot reliably execute prescribed persona shifts at predetermined points in an interaction.

**Human and AI Counterpart Effects**

As AI agents take on increasingly consequential negotiation roles—for example in procurement, dispute resolution, customer service—the field also needs to better understand whether humans respond to warmth and dominance from an AI agent as they would to these same signals from another human. The existing evidence is mixed, with some studies pointing to strong connections between human-human, human-AI, and AI-AI negotiations [50,51,52] and others highlighting important differences between the three [53,54]. Personality engineering provides a framework for systematically investigating these differences. Researchers can hold agent persona constant and vary counterpart type—human versus AI—to test whether warmth elicits the same trust and information sharing from human counterparts as it does from AI counterparts, and whether dominance provokes the same reactance.

Even when these dynamics diverge, AI-AI negotiations can remain useful for studying human negotiation if the AI-AI observations are treated as informative rather than interchangeable with human-human observations. Large AI-AI experiments can map outcome surfaces at high resolution, while smaller human-AI and human-human samples can reveal where and how those surfaces differ. Mixed-subjects designs and statistical calibration can integrate large AI-AI datasets with smaller human-AI or human-human samples to correct for systematic discrepancies across counterpart configurations [55,56]. Under appropriate assumptions, these methods can support valid inferences about human negotiation with less human data—and therefore less time and money—than human-only designs.

**Coordinate System Effects**

Negotiation research has drawn on a range of personality frameworks—the Big Five [10,57], the Dark Triad [58,59], affect models [60], and the interpersonal circumplex [14,15,30]—but these frameworks have largely developed in parallel, with each line of research generating its own distinct findings. As a result, the field lacks a clear understanding of how these coordinate systems relate to one another in the context of negotiation. Does Machiavellianism, for instance, operate through the same mechanisms as high dominance and low warmth, or does it capture something the circumplex misses? Do the Big Five dimensions that predict negotiation outcomes reduce to warmth and dominance, or do traits like conscientiousness and openness contribute independently? Without the ability to systematically compare frameworks using the same negotiation tasks and outcome measures, these questions have remained largely unanswered.

Because AI agents can be parameterized along established personality models, researchers can for the first time run otherwise identical negotiations under different coordinate systems and directly compare the resulting outcome surfaces. Now, we can test whether frameworks that appear distinct in theory produce convergent or divergent predictions in practice—and whether certain coordinate systems prove more predictive for certain outcomes or contexts than others. The interpersonal circumplex, with its parsimony and natural alignment with negotiation theory's core prescriptions [1,2,3], serves as a natural baseline, but personality engineering allows researchers to move beyond any single framework and map the relationship between them.

**Limitations**

Notwithstanding its potential, personality engineering is subject to several methodological constraints. First, the observable manifestations of warmth or dominance in an AI agent do not necessarily map onto the latent psychological structures or stable dispositions characteristic of human negotiators. Second, one should not presume that specified parameter values represent perfectly calibrated or equidistant behavioral increments: a shift from 60 to 70 in warmth, for instance, may not yield the same magnitude of change as a shift from 70 to 80, nor will it necessarily manifest consistently across distinct negotiation contexts. Finally, personality expression may be sensitive to variations in foundation models. Therefore, establishing robust and reproducible findings will consequently require empirical validation and systematic replication across diverse model configurations, prompting strategies, and human-AI contexts.

**Conclusion**

In this article, we introduce personality engineering—a methodology that uses AI agents to systematically explore the interpersonal space that negotiation theory has long identified as critical but lacked the means to empirically evaluate at scale. By introducing this approach, we hope to provide both a rigorous methodology for researchers to test the field's foundational prescriptions under controlled conditions and a practical framework for practitioners to design AI agents that negotiate effectively on behalf of people. The question of what makes a negotiator effective is decades old; AI agents offer the opportunity to develop more empirically grounded answers.

**Generative AI Statement**

During the preparation of this work, the authors used Claude for brainstorming, drafting, and revising. The authors reviewed and edited the output as needed and take full responsibility for the content of the published article.